\documentclass[10pt,twocolumn,letterpaper]{article}

\usepackage{iccv}
\usepackage{times}
\usepackage{epsfig}
\usepackage{graphicx}
\usepackage{amsmath}
\usepackage{amssymb}
\usepackage{authblk}

\usepackage[pagebackref=true,breaklinks=true,letterpaper=true,colorlinks,bookmarks=false]{hyperref}

\iccvfinalcopy 

\def\iccvPaperID{35} 

\ificcvfinal\fi

\begin{document}

\title{A weakly supervised adaptive triplet loss for deep metric learning}
\vspace{-5pt}
\author{Xiaonan Zhao\textsuperscript{1}, Huan Qi\textsuperscript{1}, Rui Luo\textsuperscript{1} and Larry Davis\textsuperscript{1,2}\\
\vspace{-5pt}
\textsuperscript{1}Amazon.com Services, Inc., Seattle, USA \textsuperscript{2}University of Maryland, College Park, USA\\
{\tt\small \{xiaonzha,qihuan,luorui,lrrydav\}@amazon.com}
}

\maketitle
\ificcvfinal \thispagestyle{empty}\fi

\begin{abstract}
   We address the problem of distance metric learning in visual similarity search, defined as learning an image embedding model which projects images into Euclidean space where semantically and visually similar images are closer and dissimilar images are further from one another. We present a weakly supervised adaptive triplet loss (ATL) capable of capturing fine-grained semantic similarity that encourages the learned image embedding models to generalize well on cross-domain data. The method uses weakly labeled product description data to implicitly determine fine grained semantic classes, avoiding the need to annotate large amounts of training data. We evaluate on the Amazon fashion retrieval benchmark and DeepFashion in-shop retrieval data. The method boosts the performance of triplet loss baseline by 10.6\% on cross-domain data and out-performs the state-of-art model on all evaluation metrics.
 
\end{abstract}

\section{Introduction}

\begin{figure}
\begin{center}
\end{center}
    \includegraphics[width=\linewidth]{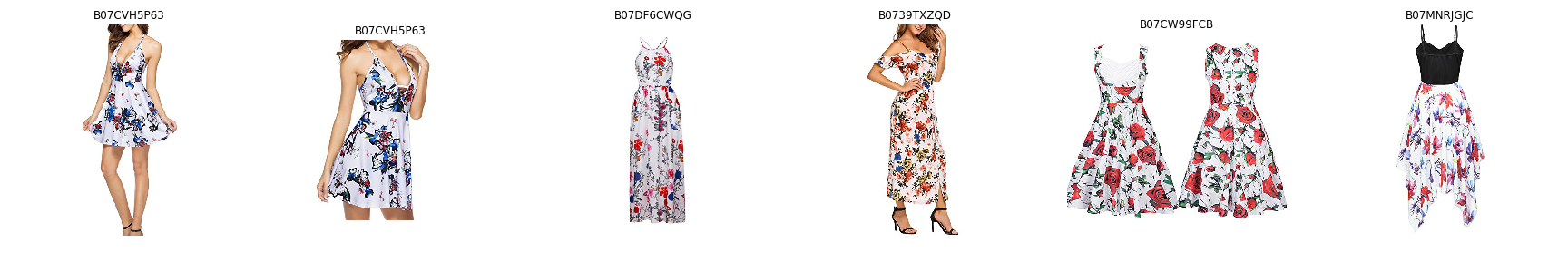}
    \includegraphics[width=\linewidth]{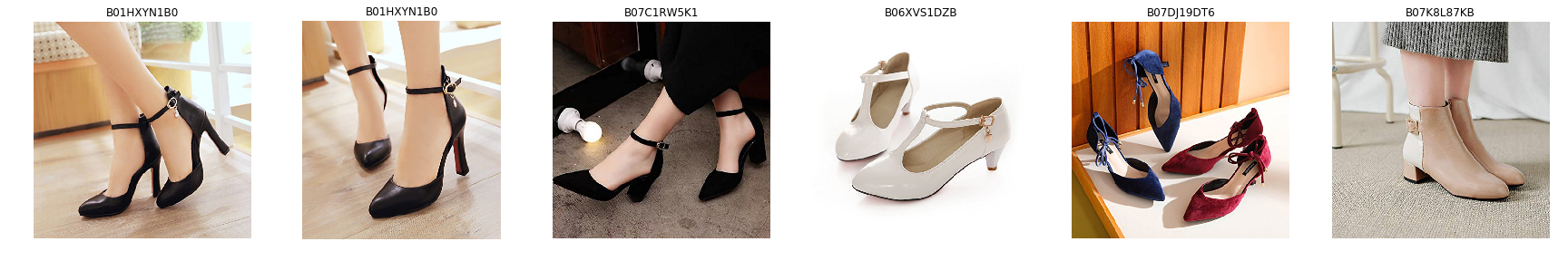}
    \includegraphics[width=\linewidth]{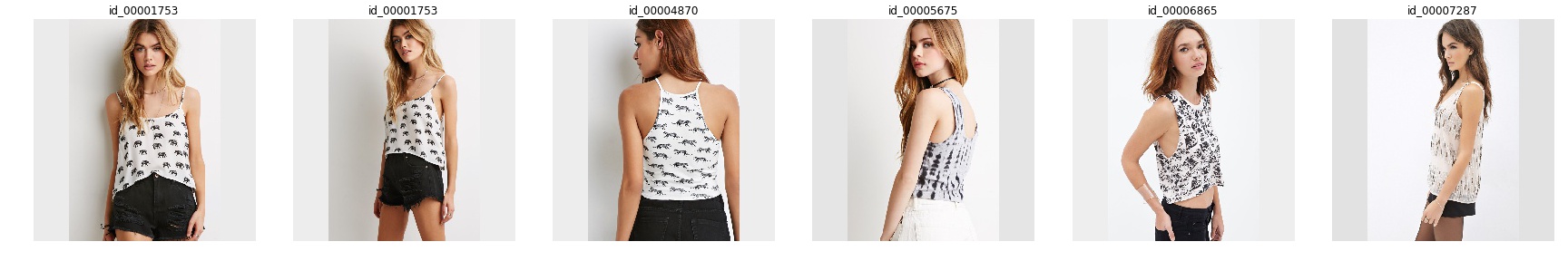}
 
   \caption{Examples of successful visually similar search results. The left most is the query image and the next 5 images are retrieved products given by our model (images belonging to the same product are merged). Product IDs are shown above the images. The first row shows one example from DeepFashion dataset, others are from Amazon dataset.}
\label{fig-sim-search}
\end{figure}

Distance metric learning is used for many computer vision tasks such as image retrieval, duplicate detection and visual search. In fashion, one important application of metric learning is visual similarity recommendation which serves customers in many ways. For example, given a query image, the model finds similar products (see Figure \ref{fig-sim-search}). Visually similar recommendations provide a customer more choices, where the core fashion features are maintained but other non-visual factors are diversified (such as price range, brand or shipping cost). 
Another important application is visual similarity based clustering, which creates a new shopping experience in addition to keyword based search. Customers can explore the whole fashion catalog through clusters which contain visually similar products. In each collection, customers are more likely to find a product which satisfies their requirements.

A wide variety of approaches have been proposed for distance metric learning such as contrastive loss \cite{hadsell2006dimensionality}, triplet loss \cite{wu2017sampling, ge2018deep} and NormSoftmax loss \cite{zhai2018making}. 
In triplet loss training, a triplet contains two images belonging to the same class, referred to as the anchor and positive samples, and a third image, from a different class, which is referred to as the negative sample. The triplet loss function is given as, $[d(a, p) - d(a, n) + m]+$, where $a$, $p$ and $n$ are anchor, positive, and negative samples, respectively. $d(\cdot, \cdot)$ is the learned metric function and $m$ is a margin term which encourages the negative sample to be further from the anchor than the positive sample. DNN based triplet loss training commonly uses stochastic gradient decent (SGD) on mini batches.

Most deep metric learning algorithms, which only use coarse-grained product ID or classes, fail to learn distances that capture fine-grained sub-categories. Such fine-grained visual similarity distances are important to learn generalized visual features and to have robust performance on cross-domain data. Here, we construct an embedding of the product from its free form text product description and use this to drive an adaptive triplet loss. This allow us to take advantage of fine-grained semantic similarity information without explicitly determining what fine-grained classes exist in our dataset or incurring the cost of labeling a large training dataset.

Our major contributions are as follows: (1) We propose a weakly supervised adaptive triplet loss (ATL) based on word embeddings of free form product descriptions. This also allows the metric learning process to generalize well on cross-domain datasets (table \ref{comp-dp}). (3) Inspired by \cite{ge2018deep}, We propose an online anchor nearest neighbor sampling (OANNS) method which addresses the batch sampling issue for large scale training data. Our method out-performs state-of-art on both our collected fashion retrieval benchmark and significantly boosts the performance on a cross-domain benchmark even without further tuning.

\section{Method}
\label{section-method}

\subsection{Data}
\label{section-weak-data}
Each product is associated with a set of images and textual description which describe the product's attributes, properties or styles. The textual descriptions can be in the form of natural language or a set of key words.
Images are relevant to the product, and no image is associated with more than one product.
The data set is denoted by $\mathcal{D} = (x_{i}, y_i, t_i), i\in [1, N]$, where $x_i$ is one of the images associated with product $y_i$. $t_i$ is the product description of $y_i$.

\subsection{Weakly supervised adaptive triplet loss}

To learn an image embedding model $\Phi(x)$ which transform an image into a unit vector, we define the mini-batch weakly supervised adaptive triplet loss (ATL) as,
\begin{equation}
    \mathcal{L_\mathcal{T}} = \sum_{(a, p, n) \in \mathcal{T}}[ d(x_a, x_p)^2 - d(x_a,x_n)^2 + \alpha(a, p, n) ]_+
    \label{equ-loss}
\end{equation}

where $(x_a, x_p, x_n)$ is a triplet of anchor image, positive image, and negative image, respectively.$\mathcal{T}$ is the mini-batch of triplets, $d(x_a, x_p) = \| \Phi(x_a) - \Phi(x_p) \|_2$ is the Euclidean distance in embedding space, the operator $[\cdot]_+$ denotes the hinge function and $\alpha(a,p, n)$ denotes the adaptive margin for the triplet in the mini-batch.

The anchor sample and positive sample are images of the same product, i.e. $y_a = y_p$ and $t_a = t_p$. Using the semantic similarity of product descriptions between anchor sample and negative sample, we define the adaptive margin $\alpha(a, p, n)$ as,
\begin{equation}
    \alpha(a,p,n) = \beta + d_{semantic}(t_a, t_n)
\end{equation}
where $d_{semantic}(t_a, t_n) \in [0, 4-\beta]$ defines the semantic similarity distance between two product description $(t_a, t_n)$. $\beta$ is the base margin term which guarantees that even for two products with the exact same product description, their margin is not zero because they are different products - i.e., we push the negative image embedding away with a minimum margin $\beta$. Because the embedding space is L2 normalized, no margin larger than 4 is possible.

We define the semantic similarity of two product description as,
\begin{equation}
    d_{semantic}(t_a, t_n) = \frac{\| g(t_a) - g(t_n) \|^2}{4 - \beta}
\end{equation}
where $g(t_a)$ is the embedding of product description $t_a$; its L2Norm is 1. Therefore, the Euclidean distance between two embedding ranges from 0 to 2. The term $4 - \beta$ is applied to satisfy the restriction that the adaptive margin lie in the range $[0, 4]$. The embedding of product description is computed simply by averaging the word vectors in the description and L2 normalizing, given as,
\begin{equation}
    g(t_a) =  \frac{\sum_{i \in [1, |t_a|]} w_i}{\|\sum_{i \in [1, |t_a|]} w_i\|_2 }
\end{equation}
where $w_i$ is the i-th word vector of product description $g(t_a)$. 

\subsection{Batch sampling and triplet construction}
The mini-batch training process is defined as,
\begin{enumerate}
	\item We sample a batch of $P$ products and then $K$ images from each product.
	\item We feed the $B = P K$ images into the image embedding model $\Phi(x)$ and get the image vectors $\mathbf{\Phi(x)} \in \mathbb{R}^{B\times d}$.
	\item From the $B$ vectors, we are able to construct $P \binom{K}{2} $ positive pairs $(x_a, x_p)$ and their embeddings $(\mathbf{\Phi(x_a)}, \mathbf{\Phi(x_p)})$.
	\item For each positive pairs, we apply the distance weighted sampling \cite{wu2017sampling} to select the negative sample $x_n$ and its embedding $\mathbf{\Phi(x_n)}$.
	\item We compute the mini-batch adaptive triplet loss and back-propagate.
\end{enumerate}

Because the batch size is limited by GPU memory in model training, it's critical to select the $P$ products that can contribute to the adaptive triplet loss. 
There are generally two families of methods for sampling the $P$ products. One is random sampling by uniform distribution\cite{wu2017sampling,zhai2018making} and the other is sampling a set of products as anchors, and then find similar products for anchor products \cite{ge2018deep}. In \cite{ge2018deep}, they compute all image vectors at the end of each epoch and use these image vectors for finding similar products in the next epoch. Computing all image vectors is a huge overhead for training, especially on a very large dataset.

To address the problem, we propose an online anchor nearest neighbor sampling (OANNS) method, which is described as,
\begin{enumerate}
	\item At the beginning of training, we assign a L2 normalized vector to each product, referred to as product vectors. These vectors are randomly initialized from a uniform distribution.
	\item For batch sampling, we first sample a batch of $N_a$ products as anchors. To fill the remaining $P-N_a$ slots, we use the most up-to-date product vectors to search for $(P/N_a - 1)$ similar products for each anchor.
	\item From the feed-forward step during batch training, we obtain the $PK$ image vectors $\mathbf{\Phi(X)}$ for the $P$ products in the current batch. Then we update the product vectors for the $P$ products in the current batch by averaging their $K$ image vectors and L2 normalizing.
\end{enumerate}

The major difference between OANNS and \cite{ge2018deep} is that we save the extra computation cost of computing image vectors for all images. During batch training, we take the image vectors from the feed-forward output and update product vectors for the current batch of products. The product vectors updated in current batch are 1 step behind the most recent model because we don't recompute the image vectors after back-propagation. In theory, we save 50\% computation cost for the feed-forwarding process. However, the freshness of similarity information is comparable to \cite{ge2018deep} - image vectors are only up-to-date in the first batch of each epoch in their training process, then become more and more out-dated until the end of the epoch.

\section{Experimental setting}
\subsection{Data}
\label{secion-data}

We collected a data set which contains 164K products, 86 product types (not mutually exclusive, e.g. apparel and dress) and 1.4 million images from amazon.com website. The training data is cleaned as follows: (1) Keep only one duplicate image for each product (2) Remove images which belong to more than one product (these images are usually not product relevant). Meta information we include in the training data are product IDs, images and the product titles. Because the training data is not quality assured by fashion specialists, it contains a number of sources of noise and ambiguity.  For example, there are images that are irrelevant to the product (e.g. image of a size chart). Product titles can be noisy. Vendors sometimes deliberately publish incorrect information in titles (e.g. vendors might put "v-neck" and "round neck" together in a title for better search indexing).

We reserved some of the data and created an Amazon Fashion catalog retrieval benchmark for evaluation. We selected the 35 most popular product types. Three fashion specialists were asked to remove irrelevant images. When we report evaluation metrics, the gallery contains all images but the query image, which is ignored in the K nearest neighbor search. The benchmark data set includes 82,465 images of 22,200 products.

We use DeepFashion in-shop retrieval benchmark \cite{liu2016deepfashion} to test the performance of models in cross-domain retrieval tasks, including 54,642 images of 11,735 clothing items. 
We followed the evaluation approach in \cite{liu2016deepfashion}. Recall@K is reported as the evaluation metric for the benchmark. 

\subsection{Implementation}
In our experiments, models are trained using MxNet \cite{Chen2015MXNetAF}. For image embedding models, ResNet50 V2 is used as the backbone. Fully connected layers after global pooling are replaced by a $2048 \times 128$ fully connected layer and an L2 Normalization layer. The final output of the model is a 128D feature vector whose L2 Norm is 1. The ResNet part is initialized with parameters which are pre-trained on ImageNet and the last fully connect layer is randomly initialized from a uniform distribution. The input images are padded to square with white background, re-sized to $224 \times 224$ and then color normalized. We used SGD with a learning rate 0.04. For those experiments using online anchor nearest neighbor sampling, $P=25$, $N_a = 5$ and $k = 4$. For experiments with adaptive triplet loss, $\beta = 0.1$. 

We implemented a baseline model for ablation study. 
The baseline model randomly samples a batch of $P = 25$ products and $k = 4$ images for each product as the mini-batch input. It uses the conventional triplet loss with a fixed margin for every triplets, $\alpha(a,p, n) = 0.1$ as described in Equation \ref{equ-loss}. 

We implemented the state-of-art algorithm NormSoftmax for comparison \cite{zhai2018making}, following the settings in the original paper. The parameters of the ResNet50 backbone are not updated in the first 6000 steps so that the added fully connected layer can be learned efficiently. The learning rate is 0.01 and the scale temperture is 0.05. 

\section{Results}

\begin{table}
 \small\addtolength{\tabcolsep}{-3pt}
  \begin{center}
  
  \centering
  \begin{tabular}{l|llllll}
    Recall@K    & 1   & 5 & 10 & 20 & 30 &  50\\ \hline
    NormSoftmax & 74.7 & 88.8 & 92.1 & 94.3 & 95.5 & 96.7 \\  \hline
    Baseline   & 63.4 & 81.1 & 86.6 & 90.6 & 92.7 & 94.5  \\ 
    +ATL    & 69.2 & 85.5 & 90.0 & 93.1 & 94.5 & 95.8  \\ 
    +OANNS   & 74.1 & 88.5 & 92.0 & 94.4 & 95.6 & 96.8  \\
    +both &  \textbf{77.0} & \textbf{90.2} & \textbf{93.2} &  \textbf{95.3} & \textbf{96.2} &  \textbf{97.2}  \\
    \hline
  \end{tabular}
  \caption{Metrics of Recall@K on DeepFashion retrieval benchmark. Weakly supervised adaptive triplet loss shows significant improvement on cross-domain benchmark. +5.8 from baseline and +2.9 from OANNS.}
  \label{comp-dp}
  \end{center}
  \vspace{-10pt}
\end{table}

\begin{table}
\small\addtolength{\tabcolsep}{-3pt}
  \begin{center}
  \centering
  \begin{tabular}{l|llllll|c}
    Recall@K    & 1   & 5 & 10 & 20 & 30 &  50 & RR@10 \\ \hline
    NormSoftmax  & 82.2   & 89.9 & 91.8 & 93.3 & 94.1 &  95.0 &  79.6 \\  \hline

    Baseline     & 78.7   & 88.1 & 90.7 & 92.7 & 93.6 &  94.4 & 75.4 \\ 
    +ATL    & 79.0  & 88.3 & 91.0 & 92.8 & 93.7 &  94.5 & 76.2 \\ 
    +OANNS     & 83.4 & 90.5 & 92.4 & 93.7 & 94.4 &  95.0 & 80.9 \\
    +both  & \textbf{83.7} & \textbf{90.8} & \textbf{92.5} & \textbf{93.9} & \textbf{94.5} & \textbf{95.3} & \textbf{81.2} \\
    \hline
  \end{tabular}
  \caption{Metrics of Recall@K and RR@10 on amazon-fashion-catalog retrieval benchmark. Columns except the last one are the numbers for recall at $K$. }
  \label{comp-afc-df}
  \end{center}
  \vspace{-10pt}
\end{table}

We evaluate the quality of image retrieval based on the widely used metric, Recall@K, following the DeepFashion benchmark evaluation\cite{ge2018deep}. We use RR@10 (or recall rate in top 10) as a measurement metric for similarity search, which is defined as the number of retrieved relevant images in the top-10 retrieval divided by the number of relevant images in the gallery. In similarity search, we want all images relevant to the query image to be ranked as top ones since they are always the ones most similar to the query image.

\vspace{-10pt}
\paragraph{Weakly supervised adaptive triplet loss}
We see an impressive performance boost from the weakly supervised adaptive triplet loss on the DeepFashion benchmark (table \ref{comp-dp}) increasing the Recall@1 from 63.4\% to 69.2\%.
Because the ATL encourages semantically dissimilar products (and related images) to be projected far away with larger margins, it encourages the model to learn features that can distinguish the dissimilar products with a large margin. By analyzing the learning curve, we found that the moodel (+ATL) can continues to converge even when the baseline model stops. This is because the ATL uses larger margins for dissimilar images which may not have violated a fixed margin.
Additionally, the loss function encourages the metric model to learn a fine-grained relative distance difference. Combined with online A-N sampling, it out-performed the NormSoftmax model on all measurement metrics -  see table \ref{comp-afc-df} and table \ref{comp-dp}.

\vspace{-10pt}
\paragraph{Online anchor nearest neighbor sampling} 
As in table \ref{comp-afc-df}, OANNS boosts the performance of the baseline model significantly. It achieves 80.9\% RR@10 on Amazon-fashion-catalog retrieval which out-performs state-of-art model NormSoftmax \cite{zhai2018making} by 1.3\%. However, the method itself is not sufficient to learn generalized image embeddings for cross-domain tasks - 74.1\% Recall@1 is slightly worse than NormSoftmax by 0.6\%. The OANNS is critical in our experiments because of its computational scaling advantage. When the model is converging, randomly sampled batches make it too easy for the model not to violate the triplet margin (i.e. zero loss).

Figure \ref{fig-sim-search} shows that our model performs well on visual similarity search. In the first row, image embeddings successfully capture some fine-grained semantic features such as "floral", "off-shoulder" and "spaghetti strap"; observe that visual features like human body, pose and number of objects are not affecting the similarity distance significantly. In the second row, given a query shoe having features such as "pointed toe", "ankle strap" and "high heel", the recommendations are shoes with all or some of the semantic features. The very different background brings a big challenge for visual similarity search.  One example from the DeepFashion dataset is shown in the third row. The distribution of images in DeepFashion dataset is quite different from our Amazon training data. In DeepFashion, there are many products on very similar looking models. The images often contain more than one product. But our model still performs very well on finding similar products for the major object in the query image.


\section{Discussion}
We have demonstrated that ATL encourages the model to learn fine-grained semantic similarity, which is critical in cross-domain visual similarity search. Our method successfully boosted performance of the triplet loss baseline model and out-performs the state-of-art metric learning model. However, there are some open questions not covered in this paper: (1) Category information such as "shoe" and "dress" contributes as much to ATL as fine-grained fashion attribute such as "strapped", "open toe", etc. This will be one of our future research topics. (2) Our model performs well on some images with complex backgrounds, however, full-body images with multiple products are still a challenge to the metric learning model.

\vspace{-10pt}
\paragraph{Acknowledgement} We thank Lailin Chen, Wei Xia, Imry Kissos, Patricia Gutierrez, Angels Borras, Etan Khanal for useful discussions, and Axel Vidales, Ben Barnes, Amy Essene, Chris Mills, Gabriel Blanco for their support.

{\small
\bibliographystyle{ieee}
\bibliography{adp_triplet_loss}
}

\end{document}